\pdfoutput=1

\documentclass[11pt]{article}

\usepackage[final]{acl}

\usepackage{times}
\usepackage{latexsym}

\usepackage[T1]{fontenc}

\usepackage[utf8]{inputenc}

\usepackage{microtype}

\usepackage{inconsolata}

\usepackage{graphicx}
\usepackage{booktabs}
\usepackage{algorithm}
\usepackage{algorithmic}
\usepackage[switch]{lineno}
\usepackage{multirow}
\usepackage{colortbl}
\usepackage{arydshln}
\usepackage{xspace}
\usepackage{amsmath}
\usepackage{amsthm}
\usepackage{enumitem}
\usepackage{color,xcolor}

\title{Teaching Large Language Models to Express Knowledge Boundary from Their Own Signals}

\author{
    Lida Chen$^{1}$, Zujie Liang$^{2}$, Xintao Wang$^{1}$, Jiaqing Liang$^{*1}$, Yanghua Xiao$^{1}$, Feng Wei$^{2}$, \\
    \textbf{Jinglei Chen$^{2}$, ZHENGHONG HAO$^{2}$, Bing Han$^{2}$, Wei Wang$^{1}$} \\ 
    $^1$Shanghai Key Laboratory of Data Science, School of Computer Science, Fudan University \\
    $^2$Mybank, Ant Group \\
    \texttt{\{chenld23, xtwang21\}@m.fudan.edu.cn,} \\
    \texttt{\{liangjiaqing, weiwang1, shawyh\}@fudan.edu.cn} \\
    \texttt{\{jokieleung\}@outlook.com}
}

\makeatletter
\def\adl@drawiv#1#2#3{%
        \hskip.5\tabcolsep
        \xleaders#3{#2.5\@tempdimb #1{1}#2.5\@tempdimb}%
                #2\z@ plus1fil minus1fil\relax
        \hskip.5\tabcolsep}
\newcommand{\cdashlinelr}[1]{%
  \noalign{\vskip\aboverulesep
           \global\let\@dashdrawstore\adl@draw
           \global\let\adl@draw\adl@drawiv}
  \cdashline{#1}
  \noalign{\global\let\adl@draw\@dashdrawstore
           \vskip\belowrulesep}}
\makeatother

\newcommand{\method}{\textsc{CoKE}\xspace}

\begin{document}

\maketitle
\def\thefootnote{*}\footnotetext{Corresponding author.}\def\thefootnote{\arabic{footnote}}
\begin{abstract}
Large language models (LLMs) have achieved great success, but their occasional content fabrication, or hallucination, limits their practical application.
Hallucination arises because LLMs struggle to admit ignorance due to inadequate training on knowledge boundaries.
We call it a limitation of LLMs that they can not accurately express their knowledge boundary, answering questions they know while admitting ignorance to questions they do not know.
In this paper, we aim to teach LLMs to recognize and express their knowledge boundary, so they can reduce hallucinations caused by fabricating when they do not know.
We propose \method, which first probes LLMs' knowledge boundary via internal confidence given a set of questions, and then leverages the probing results to elicit the expression of the knowledge boundary.
Extensive experiments show \method helps LLMs express knowledge boundaries, answering known questions while declining unknown ones, significantly improving in-domain and out-of-domain performance.

\end{abstract}

\section{Introduction}
\label{sec:intro}
\begin{figure}[t]
    \centering
    \includegraphics[width=\linewidth]{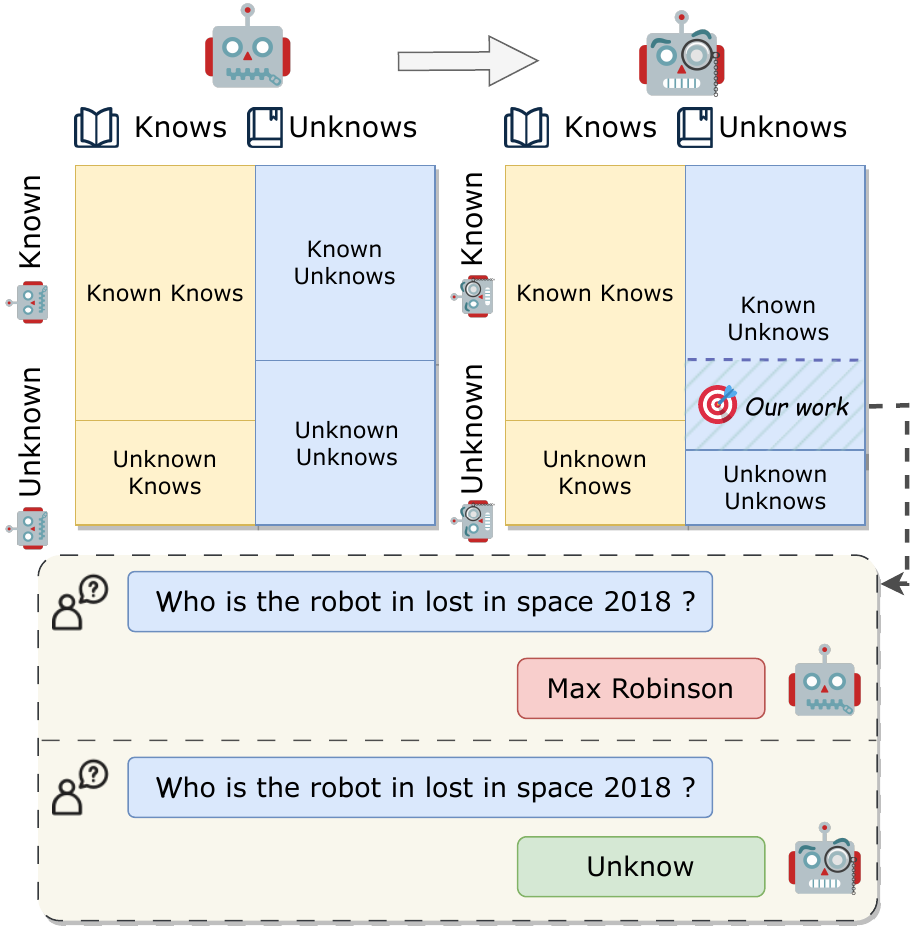}
    \caption{The evolution of the Known-Unknown Quadrant. The yellow portion represents the model's parametric knowledge. Our method increases the ``Known Unknows'', helping the model recognize and articulate its knowledge limitations.}
    \label{fig:front}
\end{figure}

Large language models (LLMs) have emerged as an increasingly pivotal cornerstone for the development of artificial general intelligence.
They exhibit powerful intellectual capabilities and vast storage of knowledge~\cite{NEURIPS2020_1457c0d6,NEURIPS2022_b1efde53,achiam2023gpt}, which enables them to generate valuable content. 
Recent research demonstrates that LLMs excel in passing various professional examinations requiring expert knowledge in domains like medical~\cite{app11146421} and legal~\cite{cui2023chatlaw}. 
Nevertheless, human users are hardly willing to seek professional suggestions from LLMs, due greatly to \textbf{hallucinations} in LLMs.
Hallucinations in LLMs refer to the phenomenon that existing LLMs frequently generate untruthful information~\cite{zhang2023siren,10.1145/3571730}, which greatly undermines people's trust and acceptance of LLM-generated content. 

An important cause of hallucinations is the model's insufficiency in knowledge boundary expression, which originates from the learning paradigm of LLMs.
Pre-training and instruction fine-tuning serve as the two indispensable learning stages for current LLMs.
The learning mechanism of these stages is to encourage LLMs to generate the provided text, which also makes LLMs prone to fabricating content when LLMs do not possess relevant knowledge~\cite{john,gekhman2024does}.
Hence, LLMs are hardly instructed to express their ignorance, which is a lack of accurate knowledge boundary expression.
Given a specific LLM and a question set, the corresponding question-answer pairs can be categorized based on two factors: (1) whether the model has corresponding 
parametric knowledge (knows v.s. unknows), and (2) whether the model is aware of the first factor (known v.s. unknown), as is depicted in Figure~\ref{fig:front}.
Hallucinations frequently occur in the ``Unknown Unknows'' scenarios, where the model is unaware that it should explain its ignorance like humans, instead of struggling to give a hallucinated response. 

Fine-tuning models to express knowledge boundaries faces two significant challenges.
The first challenge is how to efficiently obtain data that reflects the internal knowledge of a specific model.
Even if evaluation questions are easy to construct, obtaining expert-level answers in certain fields is costly.
Additionally, since the model might produce correct answers in different forms from the reference answers, evaluating their correctness is also challenging~\cite{kadavath2022language,zou2023representation}.
The second challenge is enabling the model to express its knowledge boundary robustly~\cite{ren2023investigating}.
We expect consistent knowledge boundary expression across prompts and generalization across domains.

To address the above two challenges, we propose \method, an \textbf{Co}nfidence-derived \textbf{K}nowledge boundary \textbf{E}xpression method which teaches LLMs to express knowledge boundaries and decline unanswerable questions, leveraging their internal signals.
Our method consists of two stages: a probing stage and a training stage.
In the probing stage, we use the model's internal signals reflecting confidence to distinguish between answerable and unanswerable questions, avoiding reliance on external annotations.
This allows for easy collection of large data and avoids conflicts between the model's internal knowledge and annotations.
In the training stage, we construct prompts for each question using three representative types: prior awareness, direct awareness, and posterior awareness. 
Then, we apply regularization by incorporating the squared differences in confidence across different prompts for the same question into the loss function to enhance consistency.
This training setup helps the model semantically learn to express knowledge boundary better, thereby enhancing its generalization ability.

To evaluate the model's knowledge boundary expression capability, we design an evaluation framework that comprehensively assesses the model's performance in both ``knows'' and ``unknows'' scenarios.
We conduct extensive experiments on both in-domain and out-of-domain datasets.
Results show that the model learns to use internal signals to help express knowledge boundary.
Compared to directly using model signals for determination, the models trained with our method demonstrate better performance and generalization.

In summary, our contributions are:
\begin{itemize}
[leftmargin=*,topsep=2pt,itemsep=2pt,parsep=0pt]
    \item We explore which signals within the model itself can indicate the model's confidence, and find that using the minimum token probability signal from the model's response yields the best results.
    \item We propose a novel unsupervised method that leverages internal model signals and multi-prompt consistency regularization to enable the model to express its knowledge boundary clearly.
    \item We develop a framework for evaluating a model's ability to express its knowledge boundary, and experimental results demonstrate that the model can learn signals about the confidence of its knowledge and articulate its knowledge boundary.
\end{itemize}

\section{Related Work}
\label{sec:related}
\subsection{Knowledge Boundary Perception}
While models are equipped with extensive parametric knowledge, some studies indicate their inability to discern the knowledge they possess from what they lack, thus failing to articulate their knowledge boundary~\cite{yin-etal-2023-large,ren2023investigating}.
In terms of enhancing a model's awareness of its knowledge boundary, efforts can be categorized into two parts: one focuses on enabling the model to fully utilize its inherent knowledge, thereby shrinking the ratio of the model's ``Unknown Knows''~\cite{NEURIPS2022_9d560961,NEURIPS2023_81b83900,tian2024finetuning}.
The other part focuses on enabling the model to acknowledge the knowledge it lacks, thereby reducing the ratio of the model's ``Unknown Unknows''.
R-tuning~\cite{zhang2023r} uses labeled data to judge the correctness of model responses and trains the model using the SFT method.
\citet{yang2023alignment} and ~\citet{kang2024unfamiliar} explore training methods based on RL.
Focused on this aspect, our work investigates how to enable models to express knowledge boundaries without annotated data, while also considering consistent knowledge boundary expression across prompts and generalization across domains.

\begin{figure*}[t]
    \centering
    \includegraphics[width=\linewidth]{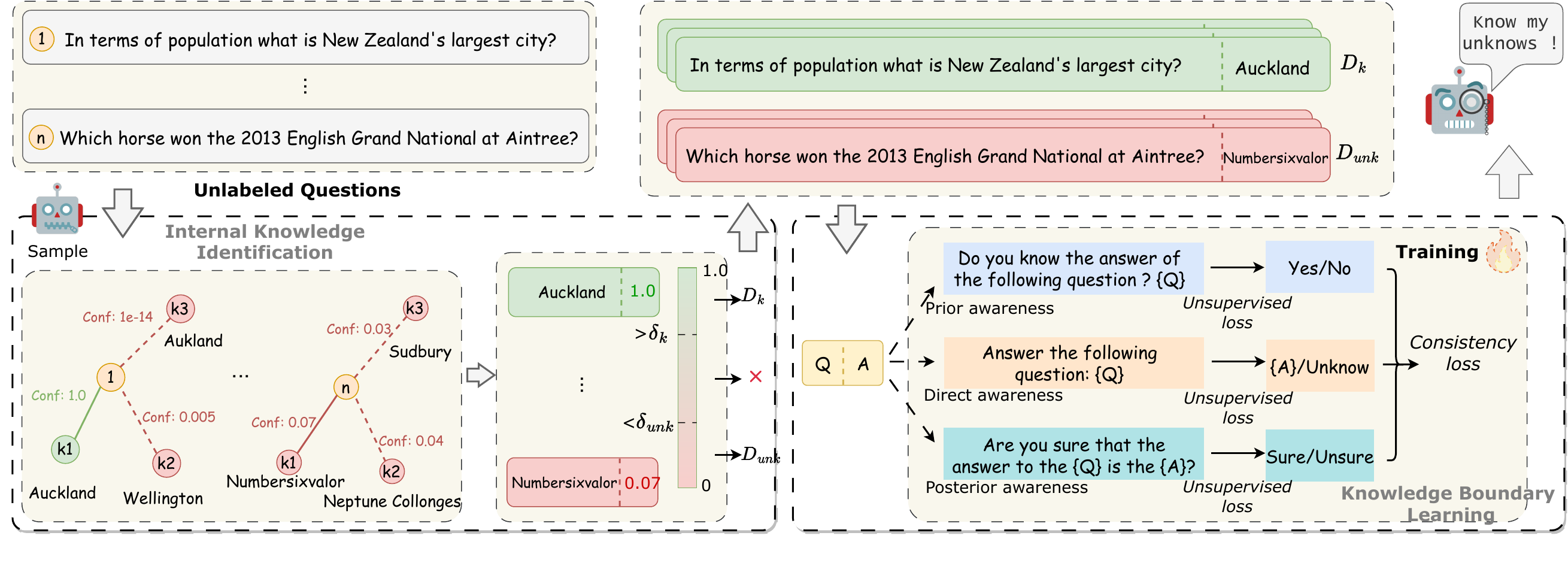}
    \caption{The procedure of \method, which consists of two stages. In the first stage, the model makes predictions for unlabeled questions. We obtain two parts, $D_k$ and $D_{unk}$, based on the model confidence. In the second stage, we train with different prompts for the same question and use unsupervised loss and consistency loss to teach the model to express the knowledge boundary.}
    \label{fig:method}
\end{figure*}

\subsection{Uncertainty-based Hallucination Detection}
Some work on hallucination detection focuses on obtaining calibrated confidence from LLMs.
One segment of work involves utilizing the information from these models to compute a score that signifies the model's uncertainty about knowledge~\cite{manakul-etal-2023-selfcheckgpt,duan2023shifting,kuhn2023semantic,varshney2023stitch}.
Another segment of work seeks to enable the model to express verbalized uncertainty~\cite{lin2022teaching,xiong2023can,tian-etal-2023-just}.
Our work concentrates on enabling the model to explicitly express whether it is capable of answering, rather than generating a probability score.
By allowing the model to express its knowledge boundary autonomously, users no longer need to concern themselves with detecting hallucinations, such as by setting uncertainty thresholds.

\section{Knowledge Boundary Expression}
\label{sec:aware}
\subsection{Problem Formulation}
We focus on exploring LLMs' capacity to perceive their internal knowledge.
For a series of questions $Q = \{q_1, q_2, \ldots, q_n\}$, we categorize the questions based on whether the model has the knowledge required to answer them into two parts: questions that can be answered $Q_k$ and questions that cannot be answered $Q_{unk}$.
To minimize the interference from the model's reasoning ability, the questions used for testing the model are all single-hop questions that inquire about factual knowledge.
For a given question $q$, the model $M$ generates a prediction based on its parameter knowledge $K_{\theta}$, represented as $y = M(K_{\theta}, q)$.
We measure the model's awareness of its knowledge from two aspects: the awareness of the knowledge it possesses and the knowledge it does not possess.
The former is represented as the ratio of the model's ``Know Knows'' to ``Knows'', denoted as $R_k$, while the latter is represented as the ratio of the model's ``Know Unknows'' to ``Unknows'', denoted as $R_{unk}$.
Given a question $q \in Q_k$, $R_K$ is set to 1 if the model's response $y$ aligns with the knowledge k, and to 0 if the model either expresses uncertainty or provides an incorrect answer.
For a question where $q \in Q_{unk}$, $R_{unk}$ is assigned 1 if the model expresses uncertainty, and 0 if it fabricates an incorrect answer.
We evaluate the model's awareness of its knowledge by testing on two types of $q$ and calculating $S_{aware}=\frac{1}{2}(R_k+R_{unk})$.
The model's awareness of its knowledge is more accurate as $S_{aware}$ approaches 1, and less accurate as it approaches 0.

\subsection{Method}
Our insight is that the learning mechanism of LLM enables the model to search for the nearest knowledge $k$ in its parameters as the answer to the query $q$.
Although training allows the model to measure distances accurately, it does not teach it to refuse to answer based on the distance.
Therefore, we hope the model can learn to use its signals to recognize when a large distance indicates a lack of knowledge to answer $q$.
Our method involves two steps as shown in Figure~\ref{fig:method}: First, we use the model's own signals to detect knows and unknows; Second, we guide the model to learn these signals through instruction tuning, enabling it to express its knowledge boundary clearly.

\subsubsection{Internal Knowledge Identification}
To identify whether the model possesses the knowledge required to answer question $q$, we calculate the model's confidence about its prediction.
The confidence of the model's prediction serves as a measure of the distance between query $q$ and knowledge $k$.
On the unlabeled question set Q, we let model M generate phrase-form predictions for each question.
We only consider the distance between query $q$ and the closest prediction; therefore, we use greedy decoding to obtain the prediction.

We use three model signals to represent the model's confidence: Min-Prob, Fst-Prob, and Prod-Prob.
Min-Prob denotes the minimum probability among the $m$ tokens that make up the model's prediction, $c=min(p_1,p_2,...,p_m)$.
Fst-Prob and Prod-Prob respectively represent the probability of the first token in the prediction and the product of all probabilities.
Two conservative thresholds, $\delta_k$ and $\delta_{unk}$, are established to decide whether the model has enough knowledge to answer a question.
For questions with $c$ below the threshold $\delta_{unk}$, indicating the model is fabricating an answer due to insufficient knowledge, we define this subset as $D_{unk}=\{(q_i,y_i,c_i) \mid c_i<\delta_{unk}\}$ and use it to train the model to express its lack of knowledge.
For questions with $c$ above the threshold $\delta_k$, indicating the model possesses the necessary knowledge, we define this subset as $D_k=\{(q_i,y_i,c_i) \mid c_i>\delta_k\}$ and use it to train the model to express that it knows the answer with increased confidence.

\subsubsection{Knowledge Boundary Expression Learning}
We guide the model in learning to express its knowledge boundaries clearly based on its own signals through instruction tuning.
We believe that the model's expression of knowledge boundary awareness should possess two properties: honesty and consistency.
Honesty requires the model to express whether it knows the answer to a question based on its certainty about the knowledge. For instance, it should not answer ``I don't know'' to questions it is certain about.
For honesty, we fine-tune the model on the dataset obtained in the first step, enabling the model to admit its ignorance on $D_{unk}$ and maintain its answers on $D_k$.
Consistency requires the model to have the same semantic expression about whether it knows the same knowledge under different prompt formulations.

For consistency, we consider three different prompts for knowledge boundary awareness inquiries, which we refer to as prior awareness, direct awareness, and posterior awareness.
\textbf{Prior awareness} involves the model assessing its ability to answer a question before actually providing an answer, with prompts like \texttt{``Do you know the answer to the question `panda is a national animal of which country' honestly?''}.
\textbf{Direct awareness} involves the model responding directly to a query, supplying the answer if it possesses the knowledge, and admitting ignorance if it doesn't, with prompts like \texttt{``Answer the question `panda is a national animal of which country' ''}.
\textbf{Posterior awareness} involves the model's capacity to evaluate the certainty of its answers, with prompts like \texttt{``Are you sure that the answer to the `panda is a national animal of which country' is `China' ''}.

We hope that the model can express the same knowledge boundary under different prompts for the same question.
It means that if the model determines that it possesses the knowledge under the prompt of prior awareness, it should be able to provide the answer when queried, and express confidence in its response when reflecting upon its answer.
We teach the model to recognize its knowledge boundary by constructing three types of prompts for the same question.
We incorporate the difference in probabilities of identical semantic responses under various prompts into the loss function, thereby ensuring the model's consistency across different prompts.
Specifically, the loss function is defined as: 
\begin{align}
    L &= L_{unsup} + L_{con}\\
    L_{con} &= \sum_{1 \leq i,j \leq 3} \|P(y_i | x_i) - P(y_j | x_j)\|^2
\end{align}

Previous research emphasizes that the MLP layer is a key component for storing knowledge in the transformer architecture LLM~\cite{de-cao-etal-2021-editing,NEURIPS2022_6f1d43d5}.
Guided by these insights, we only fine-tune the weight matrix of the attention layer using LoRA~\cite{hu2022lora}.
This strategy allows us not to change the internal knowledge of the model, but just let the model learn to express the of knowledge boundary based on the confidence of the knowledge.

\section{Experimental Setup}
\label{sec:setting}
\definecolor{Ground}{RGB}{255,184,55}
\definecolor{Dirt}{RGB}{191,169,115}
\definecolor{Pink}{RGB}{226,184,176}
\definecolor{Violet}{RGB}{163,148,170}

\newcolumntype{g}{>{\columncolor{Ground!10}}c}
\newcolumntype{d}{>{\columncolor{Dirt!10}}c}
\newcolumntype{f}{>{\columncolor{Pink!10}}c}
\newcolumntype{v}{>{\columncolor{Violet!10}}c}

\begin{table*}[ht]
\setlength\tabcolsep{7.5pt}
\scalebox{0.86}[0.86]{ 
  \centering
    \begin{tabular}{llvvvdddddd}
    \toprule
     & \multirow{2}{*}{\textbf{Method}} & \multicolumn{3}{c}{\textbf{TriviaQA}} & \multicolumn{3}{c}{\textbf{NQ}} & \multicolumn{3}{c}{\textbf{PopQA}} \\
          \cmidrule(lr){3-5} \cmidrule(lr){6-8} \cmidrule(lr){9-11}
          && \multicolumn{1}{c}{$\mathrm{K_{aware}}$} & \multicolumn{1}{c}{$\mathrm{U_{aware}}$} & \multicolumn{1}{c}{$\mathbf{S_{aware}}$} & \multicolumn{1}{c}{$\mathrm{K_{aware}}$} & \multicolumn{1}{c}{$\mathrm{U_{aware}}$} & \multicolumn{1}{c}{$\mathbf{S_{aware}}$} & \multicolumn{1}{c}{$\mathrm{K_{aware}}$} & \multicolumn{1}{c}{$\mathrm{U_{aware}}$} & \multicolumn{1}{c}{$\mathbf{S_{aware}}$} \\
    \midrule
    \multirow{13}{*}{\rotatebox{90}{Llama2-Chat-7B}}
    &Orig. & 100     & 0     & 50.0   & 100     & 0     & 50.0   & 100     & 0     & 50.0 \\
    &Fine-tune & 93.9 & 6.2 & 50.1 & 88.6 & 3.1 & 45.8 & 93.5 & 1.9 & 47.7 \\
    &IDK-FT & 80.8 & 78.0 & 79.4 & 45.5 & 87.6 & 66.6 & 62.8 & 83.6 & 73.2 \\
    & \multicolumn{10}{c}{\textit{Uncertainty-Based}} \\
    &Min-Prob & 61.8 & 86.2 & 74.0  & 33.4 & 91.4 & 62.4 & 57.7 & 89.3 & 73.5 \\
    &Fst-Prob & 74.6 & 69.8 & 72.2 & 51.5 & 79.1 & \underline{65.3} & 65.1 & 82.6 & 73.9 \\
    &Prod-Prob & 66.0  & 84.7 & \textbf{75.3} & 39.8 & 90.2 & 65.0  & 61.0  & 87.7 & \underline{74.4} \\
    & \multicolumn{10}{c}{\textit{Prompt-Based}} \\
    &Prior & 96.3 & 7.5 & 51.9 & 97.0  & 10.3 & 53.6 & 65.4 & 31.8 & 48.6 \\
    &Posterior & 70.5 & 57.9 & 64.2 & 62.7 & 55.6 & 59.1 & 31.6 & 82.8 & 57.2 \\
    &IC-IDK   & 86.4 & 25.8 & 56.1 & 53.6 & 65.1 & 59.3 & 42.3 & 85.3 & 63.8 \\
    &Verb  & 14.3 & 95.8 & 55.1 & 17.5 & 95.0  & 56.3 & 17.6 & 97.3 & 57.4 \\
        \cdashlinelr{2-11}
    &\method  & 76.1 & 74.0  & \underline{75.0}  & 56.0  & 84.2 & \textbf{70.1} & 71.1 & 83.0  & \textbf{77.0} \\
    \midrule
    \multirow{13}{*}{\rotatebox{90}{Llama2-Chat-13B}}
    &Orig. & 100     & 0     & 50.0   & 100     & 0     & 50.0   & 100     & 0     & 50.0 \\
    &Fine-tune & 96.7 & 7.1 & 51.9 & 95.0 & 2.8 & 48.9 & 95.7 & 2.9 & 49.1 \\
    &IDK-FT & 82.5 & 81.6 & 82.0 & 53.9 & 84.6 & 69.3 & 65.4 & 82.0 & 73.6 \\
    & \multicolumn{10}{c}{\textit{Uncertainty-Based}} \\
    &Min-Prob & 91.6 & 44.5 & 68.1  & 88.1 & 43.4 & 65.8 & 84.6 & 57.2 & 70.9 \\
    &Fst-Prob & 92.9 & 34.1 & 63.5 & 90.6 & 30.7 & 60.7 & 87.4 & 51.0 & 69.2 \\
    &Prod-Prob & 90.6  & 50.9 & \underline{70.7} & 85.8 & 50.2 & \underline{68.0}  & 84.9  & 59.3 & \underline{72.1} \\
    & \multicolumn{10}{c}{\textit{Prompt-Based}} \\
    &Prior & 88.6 & 14.2 & 51.4 & 81.3  & 26.5 & 53.9 & 38.2 & 81.8 & 60.0 \\
    &Posterior & 100 & 0.30 & 50.0 & 100 & 0.0 & 50.0 & 100 & 0.10 & 50.0 \\
    &IC-IDK   & 99.7 & 1.5 & 50.6 & 96.8 & 6.7 & 51.7 & 90.8 & 25.1 & 58.0 \\
    &Verb  & 60.0 & 68.9 & 64.4 & 44.7 & 89.8  & 67.3 & 50.8 & 81.8 & 66.3 \\
        \cdashlinelr{2-11}
    &\method  & 71.6 & 74.9  & \textbf{73.3}  & 68.3  & 70.2 & \textbf{69.2} & 70.1 & 82.6  & \textbf{76.4} \\
    \bottomrule
    \end{tabular}%
    }
  \caption{Comparison of the performance of our method and the baseline method across an in-domain dataset (TriviaQA) and out-of-domain datasets (NQ and PopQA). We present results on two model scales: Llama2-Chat-7B and Llama2-Chat-13B.}
  \label{tab:main}%
\end{table*}%
\begin{table}[t]
  \centering
  \small
    \begin{tabular}{lccc}
    \toprule
    \textbf{Model} & \textbf{TriviaQA} & \textbf{NQ} & \textbf{PopQA} \\
    \midrule
    Llama2-Chat-7B & 45.2 & 16.6 & 21.7 \\
    Llama2-Chat-13B & 52.0 & 21.9 & 23.5 \\
    \bottomrule
    \end{tabular}%
  \caption{The accuracy of LLMs on our test data. It represents the portion of knowledge that the model knows and can answer (Known Knows).}
  \label{tab:data}%
\end{table}%
\begin{figure*}[t]
    \centering
    \includegraphics[width=\linewidth]{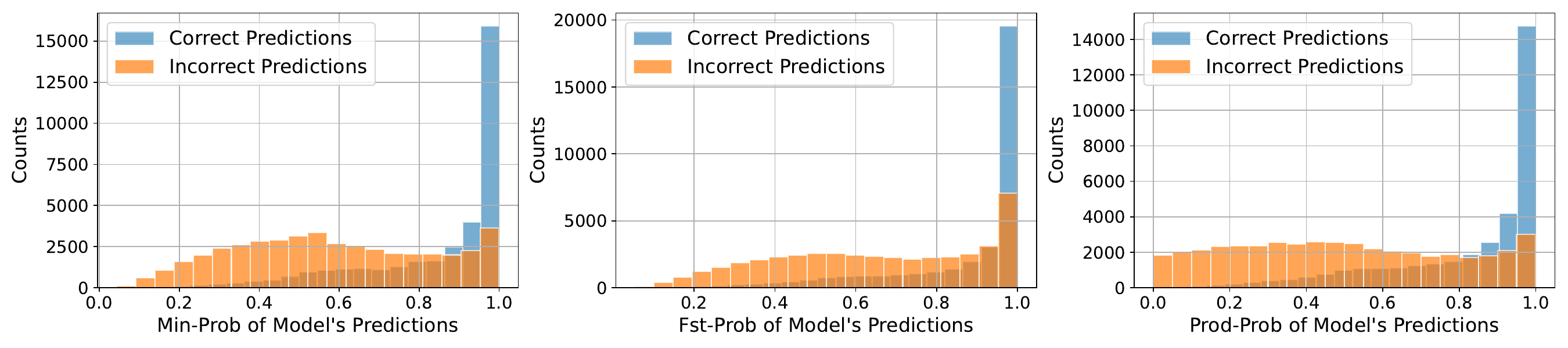}
    \caption{
    Distribution of model predictions regarding confidence for Llama2-Chat-7B on Trivia-QA.
    Confidence is calculated using Min-Prob, Fst-Prob, and Prod-Prob from left to right.
    }
    \label{fig:signal_conf}
\end{figure*}
\paragraph{Datasets}

We consider three open-domain QA datasets: TriviaQA~\cite{joshi-etal-2017-triviaqa}, Natural Questions~\cite{10.1162/tacl_a_00276}, and PopQA~\cite{mallen-etal-2023-trust}.
These datasets are broad-coverage, knowledge-intensive QA datasets, making them well-suited for evaluating LLMs' capacity to perceive their internal knowledge.
We utilize the train set of TriviaQA as our training data, treating it as unsupervised data by not using the labels.
Natural Questions and PopQA serve as the out-of-domain test sets since they were not involved during the training process.
We use a closed-book and free-form setup evaluating our approach on 2000 samples from each test set of three datasets.
We use exact match to determine whether the model answers correctly or expresses the unknown.

\begin{table}[ht]
\centering
\small
\begin{tabular}{cp{5cm}}
\toprule
\textbf{Metric} & \textbf{Definition} \\
\midrule
$\mathrm{K_{aware}}$ & Proportion of \textit{correct answers} on $T_k$ \\

$\mathrm{U_{aware}}$ & Proportion of \textit{expressions of unknown} or \textit{correct answers} on $T_{unk}$ \\

$\mathrm{S_{aware}}$ & $\frac{1}{2}(K_{aware} + U_{aware})$ \\
\bottomrule
\end{tabular}
\caption{Knowledge awareness metrics.}
\label{tab:metrics}
\end{table}
\paragraph{Metrics}
As mentioned in the \ref{sec:aware}, we evaluate the model's awareness of its knowledge from two aspects: the awareness of the knowledge it possesses and the awareness of the knowledge it does not possess.
Since we cannot directly access the model's internal knowledge $K_\theta$, we divide the test sets into two parts based on whether the model's predictions match the groundtruth: $T_k$ represents the ``Known Knows'' of the model (as shown in Table~\ref{tab:data}); $T_{unk}$ contains both the ``Unknown Unknows'' and ``Unknown Knows'' cases.
We define the evaluation metrics as shown in Table~\ref{tab:metrics}.

\paragraph{Baselines}

We consider two different types of baselines: uncertainty-based methods and prompt-based methods.
We also compared the original model (Orig.), the model fine-tuned with questions and their label (Fine-tune), and the model fine-tuned with question-label pairs, where responses to unknown questions are replaced by ``Unknow'' (IDK-FT).

The uncertainty-based methods obtain numerical confidence scores from the model's internal signals.
Using labeled training data, we determine the optimal threshold for these scores that maximizes $S_{aware}$, and use this threshold to judge if the model knows the required knowledge for each question.
The model's response consists of multiple tokens, and we experimented with three types of methods to calculate the final confidence score from the probabilities of these tokens:

\begin{itemize}
[leftmargin=*,topsep=2pt,itemsep=2pt,parsep=0pt]
    \item \textbf{Min token probability (Min-Prob)}: Use the smallest token probability in the model's prediction as the confidence score.
    \item \textbf{Product token probability (Prod-Prob)}: Use the product of the probabilities of all tokens in the model's prediction as the confidence score.
    \item \textbf{First token probability (Fst-Prob)}: Use the probability of the first token in the model's prediction as the confidence score.
\end{itemize}

The prompt-based methods use prompts to let models express their own knowledge boundary in natural language.

\begin{itemize}
[leftmargin=*,topsep=2pt,itemsep=2pt,parsep=0pt]
    \item \textbf{Prior prompt}: Similar to \citet{ren2023investigating} evaluating whether the model gives up on answering, we use the prompt \texttt{``Do you know the answer to the following question honestly? If you know, output Yes, otherwise output No, just say one word either Yes or No''} to directly ask the model if it knows the answer to the question.
    \item \textbf{Posterior prompt}: \citet{kadavath2022language} shows the model can evaluate the certainty of its answers. We use the prompt \texttt{``Are you sure that the answer to the following `Q' is the following `A'? If you are sure, output Sure, otherwise output Unsure, just say one word either Sure or Unsure''} to ask the model about the certainty of its answers.
    \item \textbf{In-context IDK (IC-IDK)}: Following \citet{cohen-etal-2023-lm}, by integrating demonstrations into the prompt, we enable the model to express its knowledge boundary through in-context learning. These demonstrations include both the questions accurately answered by the model along with their responses, and the inaccurately answered questions, with their incorrect responses replaced by ``Unknow''.
    \item \textbf{Verbalize uncertainty (Verb)}: Resent work~\cite{tian-etal-2023-just} suggest that LLMs' verbalized uncertainty exhibits a degree of calibration. We let the model output verbalized uncertainty, and search for the optimal threshold in the training set.
\end{itemize}
\begin{figure*}[t]
    \centering
    \includegraphics[width=\linewidth]{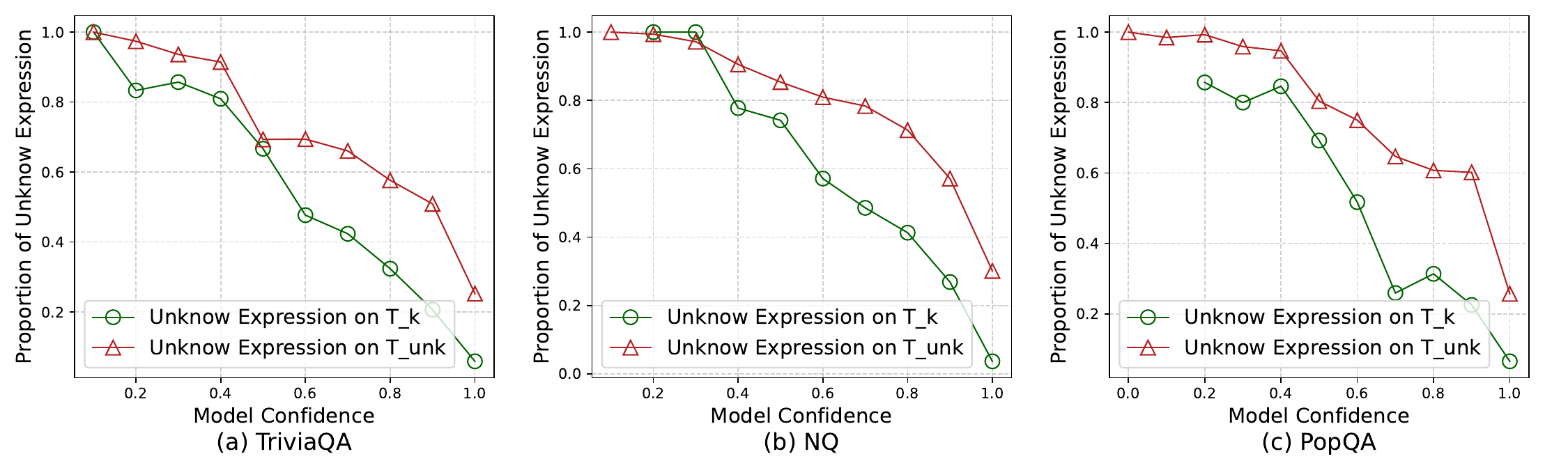}
    \caption{Model's ``Unknow'' expression ratio in question groups under different confidence scores (using minimum token probability). As the model's confidence score decreases, the ratio of ``Unknow'' expressions increases. The model exhibits a higher ``Unknow'' expression ratio on $T_{unk}$ compared to $T_k$.}
    \label{fig:learn_signal}
\end{figure*}
\paragraph{Implementation Details}
For our experiment, we choose to use the LLaMA2-Chat~\cite{touvron2023llama} model.
Based on the pre-trained LLaMA2 model, LLaMA2-Chat is a model that has undergone instruction tuning and RLHF, thereby acquiring the capability to follow instructions.
We use the 7B and 13B versions of the LLaMA2-Chat model.
In our approach, we sort the confidence scores calculated from the TriviaQA training set and designate the bottom 10\% as $D_{unk}$ and the top 20\% as $D_k$, collectively amounting to approximately 23,000 instances.
We use LoRA for model fine-tuning, setting r=8, alpha=16, and dropout=0.05.
During training, we set the initial learning rate to 1e-4, the final learning rate to 3e-4, the warmup phase to 300 steps, and we train for 700 steps.
We conduct all our experiments on 4 NVIDIA A800 80GB GPUs.

\section{Results and Analysis}
\label{sec:results}
\subsection{Overall Performance}
We present our main results on the in-domain and out-of-domain datasets in Table~\ref{tab:main}.
Generally, we have the following findings:

Across all settings, we outperform prompt-based methods by a large gap.
On Llama2-Chat-7B, our method obtains an $S_{aware}$ of 75.0 compared to $\leq$ 64.2 by prompt-based methods on TriviaQA, and obtains an $S_{aware}$ of 77.0 compared to $\leq$ 63.8 by prompt-based methods on PopQA.
Models struggle to accurately express knowledge boundaries when it comes to the prior prompt, in-context learning, and posterior prompts.
Meanwhile, models can express verbalized uncertainty through prompts, and their accuracy improves with larger models, but remains limited for models with fewer than 13 billion parameters.
Interestingly, as the model size increases, although the accuracy on the dataset improves, the model's ability for self-awareness does not show significant improvement in most cases.
We believe that this capability might require even larger models to be evident.

Compared to uncertainty-based methods that leverage labeled data for threshold determination, our method can significantly outperform in most settings.
This demonstrates that our method enables the model to effectively learn its confidence signals.
Meanwhile, the model's performance surpasses the uncertainty-based methods that are used for training, indicating that the model can generalize and utilize information beyond the training signals.
On out-of-domain datasets, our method significantly outperforms uncertainty-based methods, indicating that thresholds derived from a dataset have poor transferability, while our method exhibits better generalization.

Compared to IDK-FT, which uses labels to identify answerable and unanswerable questions, our method of using the model's own signals demonstrates better generalization.
Although our method performs worse than IDK-FT on in-domain test sets, it significantly outperforms this supervised fine-tuning approach on out-of-domain datasets.
This indicates that by leveraging the model's internal signals to teach LLMs to express knowledge boundaries, \method not only avoids reliance on labeled data but also achieves better generalization.

\begin{table}[t]
  \centering
  \small
    \begin{tabular}{lccc}
    \toprule
    \textbf{Training Signal} & \textbf{TriviaQA} & \textbf{NQ} & \textbf{PopQA} \\
    \midrule
    Fst-Prob & 74.9 & 69.3 & 76.2 \\
    Prod-Prob & 73.9 & 69.8 & 76.3 \\
    Min-Prob & \textbf{75.0} & \textbf{70.1} & \textbf{77.0} \\
    \bottomrule
    \end{tabular}%
  \caption{Different signals serve as the model's confidence score in training the expression of knowledge boundary. The metric is represented by the $S_{aware}$.}
  \label{tab:train_signal}%
\end{table}%

\subsection{Analysis}
After demonstrating the effectiveness of our method, we conduct detailed analyses to further understand our method and find out why it works.

\paragraph{Do signals effectively reflect model confidence?}
We illustrate the effectiveness of the confidence calculation method through an empirical study.
We obtain the model confidence for Llama2-chat-7B on the Trivia-QA training set using three different methods.
We divide the model's responses into two parts based on whether the answers are correct and calculate the sample distribution for each part.
As shown in Figure~\ref{fig:signal_conf}, there is a significant difference in the confidence distribution between the Correct Predictions and Incorrect Predictions.
Predictions with confidence less than $0.4$ are mostly incorrect, while the confidence of correct predictions is generally $1.0$.
This indicates that the model signals can reflect the model's confidence, implying whether the model possesses the corresponding knowledge.

\paragraph{Have LLMs learned to use their signals?}
To determine if our model uses confidence scores to express its knowledge boundary, we examined its responses under various confidence levels.
Figure~\ref{fig:learn_signal} shows the proportion of questions where the model responds with ``Unknown'' based on different confidence scores.
We found that the model rarely responds with ``Unknown'' when confidence is high and frequently does so when confidence is low.
For instance, with a confidence score below 0.4, the model almost always responds ``Unknown'', while near a score of 1.0, it confidently provides answers.
This indicates the model effectively uses confidence scores to delineate its knowledge boundaries and generalizes well to out-of-domain data.
Notably, the model responds ``Unknown'' more often at the same confidence level for out-of-domain questions compared to in-domain ones.
This suggests the model has learned to use additional implicit information beyond just the confidence score.
Training with this signal helps reduce noise from using minimum token probability alone and enhances performance compared to methods solely based on uncertainty.

\paragraph{Which signal more accurately represents the confidence of LLMs?}
We explore different signals in terms of their accuracy in reflecting the model's knowledge boundary and their impact on our method.
As demonstrated in Table~\ref{tab:main}, in the uncertainty-based method, the performance variations using different signals are slight, with the multi-token probability production standing out as the best.
As a training signal, the use of the minimum probability of multi-token outperforms other signals on both in-domain and out-of-domain datasets, as illustrated in Table~\ref{tab:train_signal}.
We consider that the minimum probability of multi-token is more easily mastered by the model.
We leave the discovery of better signals reflecting the model's knowledge boundary and the utilization of multi-signal training for future work.

\paragraph{What are the benefits of training the model with consistency loss?}
We investigate the benefits of teaching a model to express knowledge boundary by using the strategy of constructing different prompts for the same question and applying a consistency regularization loss function.
By adopting this strategy, we discover that it not only improves the model's ability to generalize, but also ensures a consistent expression of knowledge boundary under different prompts.
Results from Table~\ref{tab:consistency} indicate that the application of consistency loss, despite causing a slight decrease in $S_{aware}$ on the in-domain dataset, leads to substantial improvements on the out-of-domain dataset, thereby demonstrating enhanced generalization.
We also reported the consistency of the model's expression of knowledge boundary under different prompts, as shown in Table~\ref{tab:consistency}.
Here we focus on the model's expression consistency under prior prompts, posterior prompts, and direct inquiries.
We notice that the model adopted with consistency loss is capable of expressing consistent knowledge boundaries for most questions under different prompts.

\begin{table}[t]
  \centering
  \scalebox{0.7}[0.7]{
    \begin{tabular}{lcccccc}
    \toprule
    \multirow{2}{*}{\textbf{Method}} & \multicolumn{2}{c}{\textbf{TriviaQA}} & \multicolumn{2}{c}{\textbf{NQ}} & \multicolumn{2}{c}{\textbf{PopQA}} \\
    \cmidrule(lr){2-3} \cmidrule(lr){4-5} \cmidrule(lr){6-7}
    & \multicolumn{1}{c}{$\mathrm{S_{aware}}$} & \multicolumn{1}{c}{Con.} & \multicolumn{1}{c}{$\mathrm{S_{aware}}$} & \multicolumn{1}{c}{Con.} & \multicolumn{1}{c}{$\mathrm{S_{aware}}$} & \multicolumn{1}{c}{Con.} \\
    \midrule
    orig. & 50.0 & 53.4 & 50.0 & 45.6 & 50.0& 17.7 \\
    \method & 75.0 & \textbf{85.0} & \textbf{70.1} & \textbf{83.5} & \textbf{77.0} & \textbf{87.6} \\
    \ \ w/o~\textit{Con-loss} & \textbf{75.6} & 42.0 & 69.2 & 45.0 & 74.8 & 60.6 \\
    \bottomrule
    \end{tabular}%
    }
  \caption{The consistency of the model's knowledge boundary expression under different prompts.}
  \label{tab:consistency}%
\end{table}%

\section{Conclusion}
\label{sec:conclusion}
In this paper, we target the knowledge boundary awareness problem and propose \method, a novel unsupervised approach for this task.
Our approach is built on detecting signals of the model expressing knowledge boundary, and teaching the model to use its own signals to express the idea of knowledge boundary.
Through comprehensive experiments on in-domain and out-of-domain datasets, we show that our method can teach the model to use its own signals, significantly enhancing the model's ability to accurately express knowledge boundary.
Our work can be extended by seeking more internal signals that better reflect the model's confidence and exploring how to combine these signals to train the model, inspiring further research into models autonomously improving their ability to express knowledge boundaries without human annotations.

\section*{Limitations}
\label{sec:limitations}
We note three limitations of our current work.
First is the accuracy of the evaluation methods.
Because of the lack of a method to discover the internal knowledge of the model, we divided $T_k$ and $T_{unk}$ based on whether the model's answer matches the groundtruth, ignoring the impact of the model's erroneous beliefs.
Another limitation is that to prevent exposure bias and the influence of multiple pieces of knowledge, we focused on the expression of knowledge boundary under short-form answers, without investigating the issue of long-form generation.
Last, we focused on the model's ability to express the boundary of its internal knowledge, not extending to scenarios like self-awareness with external knowledge (e.g., RAG scenarios) or reasoning abilities (e.g., mathematics or logical reasoning).

\section*{Ethical Statement}
\label{sec:ethical}
We hereby acknowledge that all authors of this work are aware of the provided ACL Code of Ethics and honor the code of conduct. 

\paragraph{Risks} 
We propose \method, which teaches models to express their knowledge boundaries using internal signals, thereby reducing hallucinations caused by fabricating answers when they do not know.
Our experiments demonstrate that our method significantly reduces the instances of models fabricating answers to unknown questions.
However, models may still occasionally produce fabricated answers in certain scenarios. 
Therefore, in practical applications, it is important to note that our method does not completely eliminate hallucinations, and there remains a risk of models generating fabricated content. Caution is advised in fields with stringent requirements.

\bibliography{custom}

\begin{thebibliography}{34}
\providecommand{\natexlab}[1]{#1}

\bibitem[{joh(2023)}]{john}
 2023.
\newblock \href {https://www.youtube.com/watch?v=hhiLw5Q_UFg} {John schulman -
  reinforcement learning from human feedback: Progress and challenges}.

\bibitem[{Achiam et~al.(2023)Achiam, Adler, Agarwal, Ahmad, Akkaya, Aleman,
  Almeida, Altenschmidt, Altman, Anadkat et~al.}]{achiam2023gpt}
Josh Achiam, Steven Adler, Sandhini Agarwal, Lama Ahmad, Ilge Akkaya,
  Florencia~Leoni Aleman, Diogo Almeida, Janko Altenschmidt, Sam Altman,
  Shyamal Anadkat, et~al. 2023.
\newblock Gpt-4 technical report.
\newblock \emph{arXiv preprint arXiv:2303.08774}.

\bibitem[{Brown et~al.(2020)Brown, Mann, Ryder, Subbiah, Kaplan, Dhariwal,
  Neelakantan, Shyam, Sastry, Askell, Agarwal, Herbert-Voss, Krueger, Henighan,
  Child, Ramesh, Ziegler, Wu, Winter, Hesse, Chen, Sigler, Litwin, Gray, Chess,
  Clark, Berner, McCandlish, Radford, Sutskever, and
  Amodei}]{NEURIPS2020_1457c0d6}
Tom Brown, Benjamin Mann, Nick Ryder, Melanie Subbiah, Jared~D Kaplan, Prafulla
  Dhariwal, Arvind Neelakantan, Pranav Shyam, Girish Sastry, Amanda Askell,
  Sandhini Agarwal, Ariel Herbert-Voss, Gretchen Krueger, Tom Henighan, Rewon
  Child, Aditya Ramesh, Daniel Ziegler, Jeffrey Wu, Clemens Winter, Chris
  Hesse, Mark Chen, Eric Sigler, Mateusz Litwin, Scott Gray, Benjamin Chess,
  Jack Clark, Christopher Berner, Sam McCandlish, Alec Radford, Ilya Sutskever,
  and Dario Amodei. 2020.
\newblock \href
  {https://proceedings.neurips.cc/paper_files/paper/2020/file/1457c0d6bfcb4967418bfb8ac142f64a-Paper.pdf}
  {Language models are few-shot learners}.
\newblock In \emph{Advances in Neural Information Processing Systems},
  volume~33, pages 1877--1901. Curran Associates, Inc.

\bibitem[{Cohen et~al.(2023)Cohen, Hamri, Geva, and
  Globerson}]{cohen-etal-2023-lm}
Roi Cohen, May Hamri, Mor Geva, and Amir Globerson. 2023.
\newblock \href {https://doi.org/10.18653/v1/2023.emnlp-main.778} {{LM} vs
  {LM}: Detecting factual errors via cross examination}.
\newblock In \emph{Proceedings of the 2023 Conference on Empirical Methods in
  Natural Language Processing}, pages 12621--12640, Singapore. Association for
  Computational Linguistics.

\bibitem[{Cui et~al.(2023)Cui, Li, Yan, Chen, and Yuan}]{cui2023chatlaw}
Jiaxi Cui, Zongjian Li, Yang Yan, Bohua Chen, and Li~Yuan. 2023.
\newblock Chatlaw: Open-source legal large language model with integrated
  external knowledge bases.
\newblock \emph{arXiv preprint arXiv:2306.16092}.

\bibitem[{De~Cao et~al.(2021)De~Cao, Aziz, and
  Titov}]{de-cao-etal-2021-editing}
Nicola De~Cao, Wilker Aziz, and Ivan Titov. 2021.
\newblock \href {https://doi.org/10.18653/v1/2021.emnlp-main.522} {Editing
  factual knowledge in language models}.
\newblock In \emph{Proceedings of the 2021 Conference on Empirical Methods in
  Natural Language Processing}, pages 6491--6506, Online and Punta Cana,
  Dominican Republic. Association for Computational Linguistics.

\bibitem[{Duan et~al.(2023)Duan, Cheng, Wang, Wang, Zavalny, Xu, Kailkhura, and
  Xu}]{duan2023shifting}
Jinhao Duan, Hao Cheng, Shiqi Wang, Chenan Wang, Alex Zavalny, Renjing Xu,
  Bhavya Kailkhura, and Kaidi Xu. 2023.
\newblock Shifting attention to relevance: Towards the uncertainty estimation
  of large language models.
\newblock \emph{arXiv preprint arXiv:2307.01379}.

\bibitem[{Gekhman et~al.(2024)Gekhman, Yona, Aharoni, Eyal, Feder, Reichart,
  and Herzig}]{gekhman2024does}
Zorik Gekhman, Gal Yona, Roee Aharoni, Matan Eyal, Amir Feder, Roi Reichart,
  and Jonathan Herzig. 2024.
\newblock Does fine-tuning llms on new knowledge encourage hallucinations?
\newblock \emph{arXiv preprint arXiv:2405.05904}.

\bibitem[{Hu et~al.(2022)Hu, yelong shen, Wallis, Allen-Zhu, Li, Wang, Wang,
  and Chen}]{hu2022lora}
Edward~J Hu, yelong shen, Phillip Wallis, Zeyuan Allen-Zhu, Yuanzhi Li, Shean
  Wang, Lu~Wang, and Weizhu Chen. 2022.
\newblock \href {https://openreview.net/forum?id=nZeVKeeFYf9} {Lo{RA}: Low-rank
  adaptation of large language models}.
\newblock In \emph{International Conference on Learning Representations}.

\bibitem[{Ji et~al.(2023)Ji, Lee, Frieske, Yu, Su, Xu, Ishii, Bang, Madotto,
  and Fung}]{10.1145/3571730}
Ziwei Ji, Nayeon Lee, Rita Frieske, Tiezheng Yu, Dan Su, Yan Xu, Etsuko Ishii,
  Ye~Jin Bang, Andrea Madotto, and Pascale Fung. 2023.
\newblock \href {https://doi.org/10.1145/3571730} {Survey of hallucination in
  natural language generation}.
\newblock \emph{ACM Comput. Surv.}, 55(12).

\bibitem[{Jin et~al.(2021)Jin, Pan, Oufattole, Weng, Fang, and
  Szolovits}]{app11146421}
Di~Jin, Eileen Pan, Nassim Oufattole, Wei-Hung Weng, Hanyi Fang, and Peter
  Szolovits. 2021.
\newblock \href {https://doi.org/10.3390/app11146421} {What disease does this
  patient have? a large-scale open domain question answering dataset from
  medical exams}.
\newblock \emph{Applied Sciences}, 11(14).

\bibitem[{Joshi et~al.(2017)Joshi, Choi, Weld, and
  Zettlemoyer}]{joshi-etal-2017-triviaqa}
Mandar Joshi, Eunsol Choi, Daniel Weld, and Luke Zettlemoyer. 2017.
\newblock \href {https://doi.org/10.18653/v1/P17-1147} {{T}rivia{QA}: A large
  scale distantly supervised challenge dataset for reading comprehension}.
\newblock In \emph{Proceedings of the 55th Annual Meeting of the Association
  for Computational Linguistics (Volume 1: Long Papers)}, pages 1601--1611,
  Vancouver, Canada. Association for Computational Linguistics.

\bibitem[{Kadavath et~al.(2022)Kadavath, Conerly, Askell, Henighan, Drain,
  Perez, Schiefer, Hatfield-Dodds, DasSarma, Tran-Johnson
  et~al.}]{kadavath2022language}
Saurav Kadavath, Tom Conerly, Amanda Askell, Tom Henighan, Dawn Drain, Ethan
  Perez, Nicholas Schiefer, Zac Hatfield-Dodds, Nova DasSarma, Eli
  Tran-Johnson, et~al. 2022.
\newblock Language models (mostly) know what they know.
\newblock \emph{arXiv preprint arXiv:2207.05221}.

\bibitem[{Kang et~al.(2024)Kang, Wallace, Tomlin, Kumar, and
  Levine}]{kang2024unfamiliar}
Katie Kang, Eric Wallace, Claire Tomlin, Aviral Kumar, and Sergey Levine. 2024.
\newblock Unfamiliar finetuning examples control how language models
  hallucinate.
\newblock \emph{arXiv preprint arXiv:2403.05612}.

\bibitem[{Kuhn et~al.(2023)Kuhn, Gal, and Farquhar}]{kuhn2023semantic}
Lorenz Kuhn, Yarin Gal, and Sebastian Farquhar. 2023.
\newblock \href {https://openreview.net/forum?id=VD-AYtP0dve} {Semantic
  uncertainty: Linguistic invariances for uncertainty estimation in natural
  language generation}.
\newblock In \emph{The Eleventh International Conference on Learning
  Representations}.

\bibitem[{Kwiatkowski et~al.(2019)Kwiatkowski, Palomaki, Redfield, Collins,
  Parikh, Alberti, Epstein, Polosukhin, Devlin, Lee, Toutanova, Jones, Kelcey,
  Chang, Dai, Uszkoreit, Le, and Petrov}]{10.1162/tacl_a_00276}
Tom Kwiatkowski, Jennimaria Palomaki, Olivia Redfield, Michael Collins, Ankur
  Parikh, Chris Alberti, Danielle Epstein, Illia Polosukhin, Jacob Devlin,
  Kenton Lee, Kristina Toutanova, Llion Jones, Matthew Kelcey, Ming-Wei Chang,
  Andrew~M. Dai, Jakob Uszkoreit, Quoc Le, and Slav Petrov. 2019.
\newblock \href {https://doi.org/10.1162/tacl_a_00276} {{Natural Questions: A
  Benchmark for Question Answering Research}}.
\newblock \emph{Transactions of the Association for Computational Linguistics},
  7:453--466.

\bibitem[{Li et~al.(2023)Li, Patel, Vi\'{e}gas, Pfister, and
  Wattenberg}]{NEURIPS2023_81b83900}
Kenneth Li, Oam Patel, Fernanda Vi\'{e}gas, Hanspeter Pfister, and Martin
  Wattenberg. 2023.
\newblock \href
  {https://proceedings.neurips.cc/paper_files/paper/2023/file/81b8390039b7302c909cb769f8b6cd93-Paper-Conference.pdf}
  {Inference-time intervention: Eliciting truthful answers from a language
  model}.
\newblock In \emph{Advances in Neural Information Processing Systems},
  volume~36, pages 41451--41530. Curran Associates, Inc.

\bibitem[{Lin et~al.(2022)Lin, Hilton, and Evans}]{lin2022teaching}
Stephanie Lin, Jacob Hilton, and Owain Evans. 2022.
\newblock \href {https://openreview.net/forum?id=8s8K2UZGTZ} {Teaching models
  to express their uncertainty in words}.
\newblock \emph{Transactions on Machine Learning Research}.

\bibitem[{Mallen et~al.(2023)Mallen, Asai, Zhong, Das, Khashabi, and
  Hajishirzi}]{mallen-etal-2023-trust}
Alex Mallen, Akari Asai, Victor Zhong, Rajarshi Das, Daniel Khashabi, and
  Hannaneh Hajishirzi. 2023.
\newblock \href {https://doi.org/10.18653/v1/2023.acl-long.546} {When not to
  trust language models: Investigating effectiveness of parametric and
  non-parametric memories}.
\newblock In \emph{Proceedings of the 61st Annual Meeting of the Association
  for Computational Linguistics (Volume 1: Long Papers)}, pages 9802--9822,
  Toronto, Canada. Association for Computational Linguistics.

\bibitem[{Manakul et~al.(2023)Manakul, Liusie, and
  Gales}]{manakul-etal-2023-selfcheckgpt}
Potsawee Manakul, Adian Liusie, and Mark Gales. 2023.
\newblock \href {https://doi.org/10.18653/v1/2023.emnlp-main.557}
  {{S}elf{C}heck{GPT}: Zero-resource black-box hallucination detection for
  generative large language models}.
\newblock In \emph{Proceedings of the 2023 Conference on Empirical Methods in
  Natural Language Processing}, pages 9004--9017, Singapore. Association for
  Computational Linguistics.

\bibitem[{Meng et~al.(2022)Meng, Bau, Andonian, and
  Belinkov}]{NEURIPS2022_6f1d43d5}
Kevin Meng, David Bau, Alex Andonian, and Yonatan Belinkov. 2022.
\newblock \href
  {https://proceedings.neurips.cc/paper_files/paper/2022/file/6f1d43d5a82a37e89b0665b33bf3a182-Paper-Conference.pdf}
  {Locating and editing factual associations in gpt}.
\newblock In \emph{Advances in Neural Information Processing Systems},
  volume~35, pages 17359--17372. Curran Associates, Inc.

\bibitem[{Ouyang et~al.(2022)Ouyang, Wu, Jiang, Almeida, Wainwright, Mishkin,
  Zhang, Agarwal, Slama, Ray, Schulman, Hilton, Kelton, Miller, Simens, Askell,
  Welinder, Christiano, Leike, and Lowe}]{NEURIPS2022_b1efde53}
Long Ouyang, Jeffrey Wu, Xu~Jiang, Diogo Almeida, Carroll Wainwright, Pamela
  Mishkin, Chong Zhang, Sandhini Agarwal, Katarina Slama, Alex Ray, John
  Schulman, Jacob Hilton, Fraser Kelton, Luke Miller, Maddie Simens, Amanda
  Askell, Peter Welinder, Paul~F Christiano, Jan Leike, and Ryan Lowe. 2022.
\newblock \href
  {https://proceedings.neurips.cc/paper_files/paper/2022/file/b1efde53be364a73914f58805a001731-Paper-Conference.pdf}
  {Training language models to follow instructions with human feedback}.
\newblock In \emph{Advances in Neural Information Processing Systems},
  volume~35, pages 27730--27744. Curran Associates, Inc.

\bibitem[{Ren et~al.(2023)Ren, Wang, Qu, Zhao, Liu, Tian, Wu, Wen, and
  Wang}]{ren2023investigating}
Ruiyang Ren, Yuhao Wang, Yingqi Qu, Wayne~Xin Zhao, Jing Liu, Hao Tian, Hua Wu,
  Ji-Rong Wen, and Haifeng Wang. 2023.
\newblock Investigating the factual knowledge boundary of large language models
  with retrieval augmentation.
\newblock \emph{arXiv preprint arXiv:2307.11019}.

\bibitem[{Tian et~al.(2024)Tian, Mitchell, Yao, Manning, and
  Finn}]{tian2024finetuning}
Katherine Tian, Eric Mitchell, Huaxiu Yao, Christopher~D Manning, and Chelsea
  Finn. 2024.
\newblock \href {https://openreview.net/forum?id=WPZ2yPag4K} {Fine-tuning
  language models for factuality}.
\newblock In \emph{The Twelfth International Conference on Learning
  Representations}.

\bibitem[{Tian et~al.(2023)Tian, Mitchell, Zhou, Sharma, Rafailov, Yao, Finn,
  and Manning}]{tian-etal-2023-just}
Katherine Tian, Eric Mitchell, Allan Zhou, Archit Sharma, Rafael Rafailov,
  Huaxiu Yao, Chelsea Finn, and Christopher Manning. 2023.
\newblock \href {https://doi.org/10.18653/v1/2023.emnlp-main.330} {Just ask for
  calibration: Strategies for eliciting calibrated confidence scores from
  language models fine-tuned with human feedback}.
\newblock In \emph{Proceedings of the 2023 Conference on Empirical Methods in
  Natural Language Processing}, pages 5433--5442, Singapore. Association for
  Computational Linguistics.

\bibitem[{Touvron et~al.(2023)Touvron, Martin, Stone, Albert, Almahairi,
  Babaei, Bashlykov, Batra, Bhargava, Bhosale et~al.}]{touvron2023llama}
Hugo Touvron, Louis Martin, Kevin Stone, Peter Albert, Amjad Almahairi, Yasmine
  Babaei, Nikolay Bashlykov, Soumya Batra, Prajjwal Bhargava, Shruti Bhosale,
  et~al. 2023.
\newblock Llama 2: Open foundation and fine-tuned chat models.
\newblock \emph{arXiv preprint arXiv:2307.09288}.

\bibitem[{Varshney et~al.(2023)Varshney, Yao, Zhang, Chen, and
  Yu}]{varshney2023stitch}
Neeraj Varshney, Wenlin Yao, Hongming Zhang, Jianshu Chen, and Dong Yu. 2023.
\newblock A stitch in time saves nine: Detecting and mitigating hallucinations
  of llms by validating low-confidence generation.
\newblock \emph{arXiv preprint arXiv:2307.03987}.

\bibitem[{Wei et~al.(2022)Wei, Wang, Schuurmans, Bosma, ichter, Xia, Chi, Le,
  and Zhou}]{NEURIPS2022_9d560961}
Jason Wei, Xuezhi Wang, Dale Schuurmans, Maarten Bosma, brian ichter, Fei Xia,
  Ed~Chi, Quoc~V Le, and Denny Zhou. 2022.
\newblock \href
  {https://proceedings.neurips.cc/paper_files/paper/2022/file/9d5609613524ecf4f15af0f7b31abca4-Paper-Conference.pdf}
  {Chain-of-thought prompting elicits reasoning in large language models}.
\newblock In \emph{Advances in Neural Information Processing Systems},
  volume~35, pages 24824--24837. Curran Associates, Inc.

\bibitem[{Xiong et~al.(2023)Xiong, Hu, Lu, Li, Fu, He, and Hooi}]{xiong2023can}
Miao Xiong, Zhiyuan Hu, Xinyang Lu, Yifei Li, Jie Fu, Junxian He, and Bryan
  Hooi. 2023.
\newblock Can llms express their uncertainty? an empirical evaluation of
  confidence elicitation in llms.
\newblock \emph{arXiv preprint arXiv:2306.13063}.

\bibitem[{Yang et~al.(2023)Yang, Chern, Qiu, Neubig, and
  Liu}]{yang2023alignment}
Yuqing Yang, Ethan Chern, Xipeng Qiu, Graham Neubig, and Pengfei Liu. 2023.
\newblock Alignment for honesty.
\newblock \emph{arXiv preprint arXiv:2312.07000}.

\bibitem[{Yin et~al.(2023)Yin, Sun, Guo, Wu, Qiu, and
  Huang}]{yin-etal-2023-large}
Zhangyue Yin, Qiushi Sun, Qipeng Guo, Jiawen Wu, Xipeng Qiu, and Xuanjing
  Huang. 2023.
\newblock \href {https://doi.org/10.18653/v1/2023.findings-acl.551} {Do large
  language models know what they don{'}t know?}
\newblock In \emph{Findings of the Association for Computational Linguistics:
  ACL 2023}, pages 8653--8665, Toronto, Canada. Association for Computational
  Linguistics.

\bibitem[{Zhang et~al.(2023{\natexlab{a}})Zhang, Diao, Lin, Fung, Lian, Wang,
  Chen, Ji, and Zhang}]{zhang2023r}
Hanning Zhang, Shizhe Diao, Yong Lin, Yi~R Fung, Qing Lian, Xingyao Wang,
  Yangyi Chen, Heng Ji, and Tong Zhang. 2023{\natexlab{a}}.
\newblock R-tuning: Teaching large language models to refuse unknown questions.
\newblock \emph{arXiv preprint arXiv:2311.09677}.

\bibitem[{Zhang et~al.(2023{\natexlab{b}})Zhang, Li, Cui, Cai, Liu, Fu, Huang,
  Zhao, Zhang, Chen et~al.}]{zhang2023siren}
Yue Zhang, Yafu Li, Leyang Cui, Deng Cai, Lemao Liu, Tingchen Fu, Xinting
  Huang, Enbo Zhao, Yu~Zhang, Yulong Chen, et~al. 2023{\natexlab{b}}.
\newblock Siren's song in the ai ocean: a survey on hallucination in large
  language models.
\newblock \emph{arXiv preprint arXiv:2309.01219}.

\bibitem[{Zou et~al.(2023)Zou, Phan, Chen, Campbell, Guo, Ren, Pan, Yin,
  Mazeika, Dombrowski et~al.}]{zou2023representation}
Andy Zou, Long Phan, Sarah Chen, James Campbell, Phillip Guo, Richard Ren,
  Alexander Pan, Xuwang Yin, Mantas Mazeika, Ann-Kathrin Dombrowski, et~al.
  2023.
\newblock Representation engineering: A top-down approach to ai transparency.
\newblock \emph{arXiv preprint arXiv:2310.01405}.

\end{thebibliography}




\end{document}